\documentclass[letterpaper, 10 pt, conference]{ieeeconf}
\IEEEoverridecommandlockouts                              % This command is only needed if 
\usepackage{times}
\usepackage{epsfig}
\usepackage{graphicx}
\usepackage{amsmath}
\usepackage{amssymb}

\usepackage{amsmath}
\usepackage{graphics}
\usepackage{graphicx}
\usepackage{amssymb}
\usepackage{algorithm}
\usepackage{algorithmic}
\usepackage{mathrsfs}
\usepackage{booktabs}
\usepackage{bm}
\usepackage{bbm}
\usepackage{float}
\usepackage{url}
\usepackage{amsfonts}
\usepackage{multirow}
\usepackage{textcomp}
\usepackage{supertabular}
\usepackage{array}
\usepackage{caption}
\usepackage{gensymb}
\usepackage{fancyhdr}
\fancypagestyle{firstpage}{%
  \lhead{This paper is accepted and presented in ICRA 2022}
}
\title{\LARGE \bf
Real-time Full-stack Traffic Scene Perception for Autonomous Driving with Roadside Cameras
}

\begin{document}

\author{Zhengxia Zou$^{1}$, Rusheng Zhang$^{1}$, Shengyin Shen$^{1}$, Gaurav Pandey $^{2}$, Punarjay Chakravarty$^{2}$, \\ Armin Parchami$^{2}$, and Henry X. Liu$^{1*}$% <-this % stops a space
\thanks{$^{1}$Zhengxia Zou, Rusheng Zhang, Shengyin Shen, and Henry X. Liu are with the Department of Civil and Environmental Engineering, University of Michigan, Ann Arbor.}
\thanks{$^{2}$Gaurav Pandey, Punarjay Chakravarty, and Armin Parchami are with Ford Motor Company.}
\thanks{$^{*}$Corresponding author, henryliu@umich.edu.}
}
\maketitle
\thispagestyle{firstpage}
\pagestyle{empty}

%%%%%%%%%%%%%%%%%%%%%%%%%%%%%%%%%%%%%%%%%%%%%%%%%%%%%%%%%%%%%%%%%%%%%%%%%%%%%%%%
\begin{abstract}

We propose a novel and pragmatic framework for traffic scene perception with roadside cameras. The proposed framework covers a full-stack of roadside perception pipeline for infrastructure-assisted autonomous driving, including object detection, object localization, object tracking, and multi-camera information fusion. Unlike previous vision-based perception frameworks rely upon depth offset or 3D annotation at training, we adopt a modular decoupling design and introduce a landmark-based 3D localization method, where the detection and localization can be well decoupled so that the model can be easily trained based on only 2D annotations. The proposed framework applies to either optical or thermal cameras with pinhole or fish-eye lenses. Our framework is deployed at a two-lane roundabout located at Ellsworth Rd. and State St., Ann Arbor, MI, USA, providing 7x24 real-time traffic flow monitoring and high-precision vehicle trajectory extraction. The whole system runs efficiently on a low-power edge computing device with all-component end-to-end delay of less than 20ms.

\end{abstract}

%%%%%%%%%%%%%%%%%%%%%%%%%%%%%%%%%%%%%%%%%%%%%%%%%%%%%%%%%%%%%%%%%%%%%%%%%%%%%%%%
\section{INTRODUCTION}

Infrastructure-assisted cooperative perception is an emerging research topic in autonomous driving and intelligent transportation. Recently, the rapid development of deep learning and computer vision technology has opened up new perspectives for assisting automated vehicles in complex driving environments. With roadside sensors, hazardous driving scenarios could be identified (e.g. objects hidden in the blind spot), and automated vehicles could be informed in advance. 

\begin{figure}
    \centering{\includegraphics[width=0.8\linewidth]{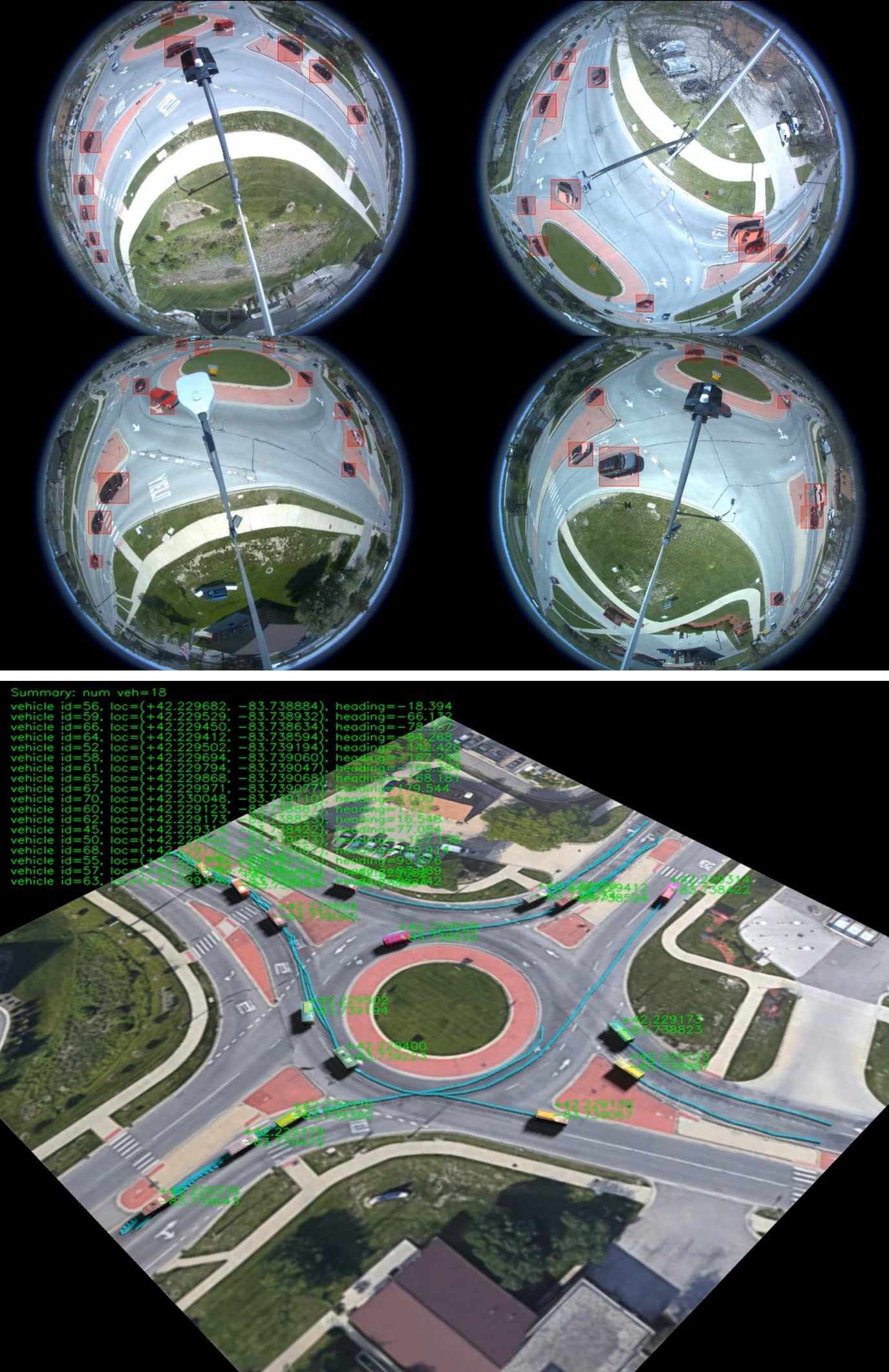}} \\
    \caption{We propose a vision-based solution for real-time traffic object detection, localization, information fusion, and tracking. The upper image shows real-time detection results with four 360\degree \ roadside fish-eye cameras and the lower one shows the vectorized location and trajectory of each object.}
    \label{fig:teaser}
\end{figure}

\begin{figure*}
    \centering{\includegraphics[width=0.9\linewidth]{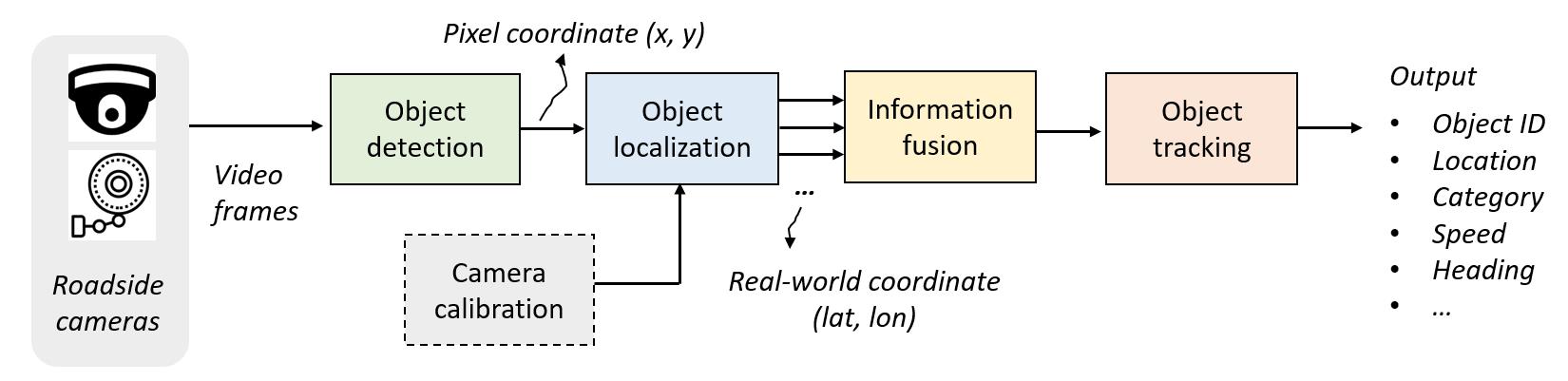}} \\
    \caption{An overview of the proposed framework for roadside vision-based traffic scene perception.}
    \label{fig:overview}
\end{figure*}

In this paper, we propose a novel and pragmatic solution for roadside camera-based perception. As shown in Fig.~\ref{fig:teaser}, the proposed scheme covers a full-stack of roadside perception pipeline for infrastructure-assisted autonomous driving - from object detection, localization, tracking, to multi-sensor information fusion. To obtain the real-world object location from a 2D image, previous 3D detection methods~\cite{liu2020smoke,manhardt2019roi,simonelli2019disentangling} typically require camera calibration parameters or depth offset available at training so that a transformation between the image plane and the 3D world can be constructed. However, such information is difficult to obtain in data annotation phase. Particularly, the calibration of camera extrinsic parameters may rely heavily on other types of sensors (such as lidar) and may also involve the issues of joint calibration and multi-sensor synchronization~\cite{geiger2012we}.

Instead of using multi-sensor joint calibration, we introduce a purely vision-based solution with a detection-localization decoupling design. In our method, a landmark-based object localization strategy is utilized that allows our detector to be trained solely based on 2D annotations. The detection results are then lifted to 3D with the landmark Homography and camera intrinsics. Our method can be applied to both optical and thermal cameras with pinhole or fish-eye lenses. Using a lightweight 
MobileNet-v2~\cite{sandler2018mobilenetv2} network backbone, our method can run efficiently in real-time on a low-power edge computing box. The all-component end-to-end perception delay is less than 20ms.

Our contributions are summarized as follows.

\begin{itemize}
\item We propose a novel framework for full-stack roadside assisted traffic scene perception, including object detection, 3D localization, tracking, and multi-camera information fusion. Our method is flexible and scalable - since the training of our model only requires 2D annotations, the whole framework can be deployed quickly and migrated elegantly at any new application scenarios.
\item Most previous perception systems for autonomous driving focus on onboard perception only and rarely discuss roadside-based solutions. To our best knowledge, we are one of the first to propose and implement a fully established roadside framework for infrastructure-assisted autonomous driving.
\item  Our framework is deployed at a two-lane roundabout in Ann Arbor, MI, providing 7x24 traffic flow monitoring and hazardous driving warnings capabilities. For the entire 10000 m$^2$ roundabout area, our method achieves sub-meter-level localization accuracy with a single camera and 0.4m localization accuracy with information fusion of multiple cameras.
\end{itemize}

\section{RELATED WORK}

Roadside sensor-based perception system has a long history and can be traced back to 1980s~\cite{datondji2016survey}. To detect traffic objects and monitor their behavior, some early methods are developed based on traditional computer vision techniques such as background subtraction~\cite{furuya2014road}, frame difference~\cite{messelodi2005computer}, optical flow~\cite{li2016robust}, etc. Recently, the fast development of deep learning technology has greatly promoted object detection and tracking research. Some representative approaches includes Faster R-CNN~\cite{girshick2014rich,girshick2015fast,ren2015faster}, SSD~\cite{liu2016ssd}, and YOLO~\cite{redmon2016you,redmon2017yolo9000,redmon2018yolov3} for object detection; DeepSort~\cite{wojke2017simple} and Center Track~\cite{zhou2020tracking} for object tracking. Some of these methods have been successfully applied to UAV-based traffic surveillance applications~\cite{zhang2017application}. However, for roadside-based traffic perception, deep learning-based approaches are still in their infancy and have attracted increasing attention recently~\cite{bai2022infrastructure}.

2D/3D object detection plays a central role in roadside traffic scene perception.
The task of 2D object detection~\cite{ren2015faster} is to find the pixel location of all objects of interest in the image and determine their bounding boxes and categories. In contrast to conventional 2D object detection, 3D object detection predicts 3D boxes (with 3D location, orientation, and size) from a single monocular image~\cite{chen2016monocular,simonelli2019disentangling,liu2020smoke,manhardt2019roi} or stereo images~\cite{li2019stereo}, which has received great attention in autonomous driving recently. The proposed detection method is mostly related to Objects as Points~\cite{zhou2019objects}, a recent popular 2D detection framework. We use a similar idea of point detection but extend this framework for 3D pose and 3D size estimation with additional output branches. Instead of predicting the center of 2D box, we predict the object's 3D bottom center and lift the prediction to 3D using a pre-calibrated plane-to-plane Homography. Compared to recent 3D object detection methods, our ``point detection + 3D lifting'' design makes our method neither requires depth information nor 3D annotation during the training,  greatly reducing the cost of data annotation and collection. In addition, most current 3D object detection solutions of autonomous driving only focus on onboard perception and rarely discuss roadside-based perception. In contrast to previous onboard solutions~\cite{chen2016monocular,simonelli2019disentangling,liu2020smoke,manhardt2019roi}, we provide a new framework for roadside-based perception and have evaluated the effectiveness of our system at a two-lane roundabout with real-world connected and automated vehicles.

\begin{figure*}
    \centering{\includegraphics[width=0.9\linewidth]{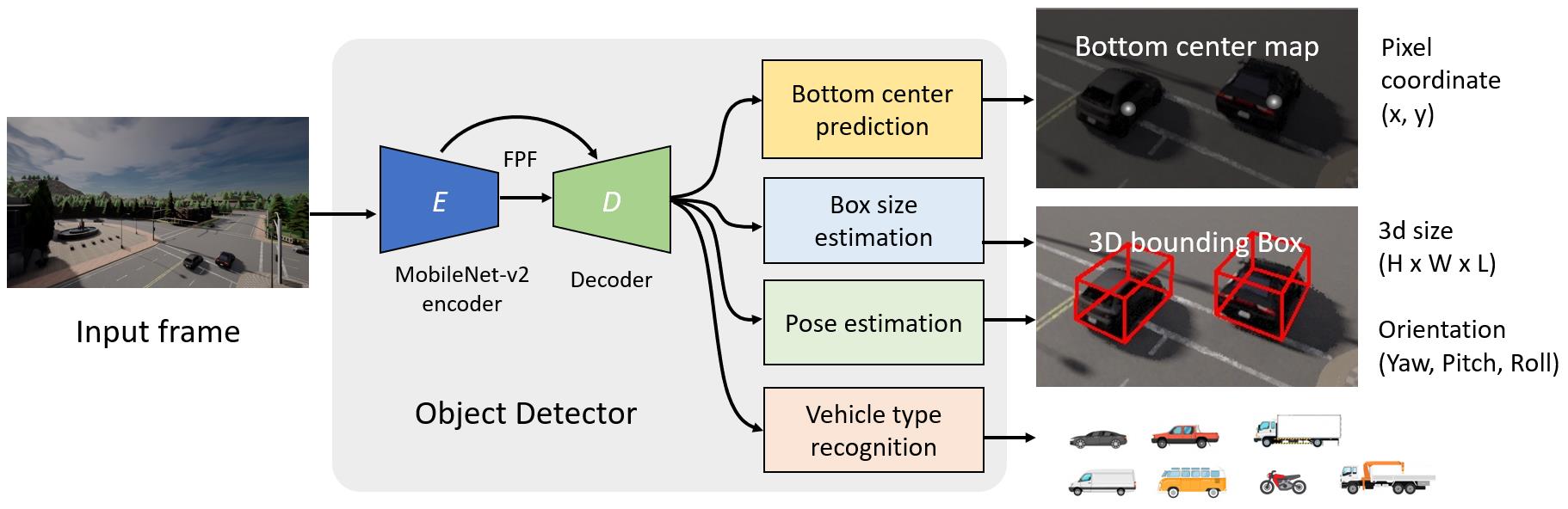}} \\
    \caption{Architecture of the proposed detection method. Our detector consists of a feature encoder~\cite{sandler2018mobilenetv2}, a feature decoder, and four output heads designed for vehicle bottom-center prediction, box-size estimation, pose estimation, vehicle type recognition.}
    \label{fig:detector}
\end{figure*}

\section{Methodology}

The introduced framework is composed of four different modules: 1. object detection, 2. object localization, 3. information fusion, and 4. object tracking. Fig.~\ref{fig:overview} shows an overview of the proposed framework. The object detection operates directly on 2D images and generates 2D bounding boxes; the object localization lifts the 2D detection to the 3D world; detections from different sensors are fused; finally, individual ids will be assigned for all detected vehicles with tracking.

\subsection{Object Detection}

A single-stage center-aware detector is designed for joint object detection, pose estimation, and category recognition. As shown in  Fig.~\ref{fig:detector}, the proposed detector consists of a lightweight image encoder $E$, a feature decoder $D$, and four prediction heads (for bottom center prediction, box-size estimation, pose-estimation, and vehicle type recognition, respectively). To improve detection on small objects, we apply feature pyramid fusion~\cite{lin2017feature} in our decoder and progressively upsample the feature map to the same spatial size as the input. In the following, we will introduce the four prediction heads accordingly.

\subsubsection{Bottom-center prediction}

The bottom-center prediction branch is trained to produce a heat-map with the same spatial size as the input. We define the loss function of the bottom-center prediction branch as a pixel-wise least-square loss between the prediction and ground truth:
\begin{equation}\label{eq:bottom-center-loss}
    \mathcal{L}_{center}(X) = \mathbb{E}_{X\sim \mathcal{D}}\{\text{TopK}(\| Y_{center} - \hat{Y}_{center} \|_2^2)\},
\end{equation}
where $Y_{center}$ and $\hat{Y}_{center}$ are the prediction output and its 2D ground truth map. $X$ and $\mathcal{D}$ are the input image and the dataset. TopK represents hard-example selection - in each training iteration, only the top 1\% of pixels with the largest loss will be used for error back-propagation. In $\hat{Y}_{center}$, a larger pixel value means a larger probability the pixel belongs to the bottom center of an object. We generate the ground truth maps with a Gaussian function:
\begin{equation}
\hat{Y}_{center}(i,j)=\sum_t^T\exp{(-d_t(i,j)^2/\sigma_t^2)},
\end{equation}
where $(i,j)$ is the pixel location; $T$ is the number of object in an image; $d_t(i,j)$ is the distance between the $(i,j)$ to the bottom center of the $t$-th object; $\sigma_t=\frac{1}{2}\sqrt{l_t}$; $l_t$ is the pixel bounding box diagonal length of the $t$-th object. 

\subsubsection{3D Size and Pose Estimation}

The 3D size prediction and pose estimation can be formulated as least square regression problems. The loss function of the 3D size branch and pose estimation branch are defined as follows:
\begin{equation}
\begin{split}
    \mathcal{L}_{size}(X) &= \mathbb{E}_{X\sim \mathcal{D}}\{\hat{Y}_{center}(\| \log Y_{size} - \log \hat{Y}_{size} \|_2^2)\},\\
    \mathcal{L}_{pose}(X) &= \mathbb{E}_{X\sim \mathcal{D}}\{\hat{Y}_{center}(\| Y_{pose} - \hat{Y}_{pose} \|_2^2)\},
\end{split}
\end{equation}
where $Y_{pose}$ and $Y_{size}$ are the predicted pose and size maps. We apply log normalization to the predicted size for better convergence. $\hat{Y}_{pose}$ and $\log \hat{Y}_{size}$ are their ground truth. We use the ground truth bottom center $\hat{Y}_{center}$ as a pixel-wise weight map since the predictions only need to be focused on the object regions.

\subsubsection{Object Category Recognition}

The vehicle type recognition can be considered as a standard classification problem. We therefore define the loss as a standard cross-entropy distance between the probabilistic output and the ground truth. The loss function is defined as follows: 
\begin{equation}
    \mathcal{L}_{v-type}(X) = \mathbb{E}_{X\sim \mathcal{D}}\{-\hat{Y}_{center}\sum_i^C \hat{Y}_{type}^{(i)} \log Y_{type}^{(i)}\},
\end{equation}
where $Y_{type}$ is the predicted category probability maps after softmax normalization; $\hat{Y}_{type}$ is the one-hot ground truth; $C$ is the number of vehicle category.

\subsubsection{Multi-task Loss} 

We finally train our detector by following multi-task loss function as follows:
\begin{equation}
    \mathcal{L} = \mathcal{L}_{center} + \beta_1\mathcal{L}_{size} + \beta_2\mathcal{L}_{pose} + \beta_3\mathcal{L}_{v-type}
\end{equation}
where $\beta_1$, $\beta_2$, and $\beta_3$ are predefined weights for balancing the loss terms from different prediction heads. Since all output branches are differentiable, we can train the whole detector in an end-to-end fashion.

\subsubsection{Network configuration} 

We use a similar network configuration in all output branches. In each output, we use a stacked two convolutional layers on top of the decoder feature map for prediction. We choose Sigmoid output activation for bottom center prediction, Tanh for normalized pose prediction, ReLU for size prediction, and Softmax for category recognition.

\subsection{Camera Calibration and Object Localization}

Since our object detector is only trained with 2D annotations, to determine their real-world location, a mapping needs to be constructed between the pixel space and the 3D world. Here we introduce a simple and elegant solution for camera calibration and object localization. Instead of estimating the intrinsic/extrinsic camera matrices jointly with other sensors, we directly transform the image into a bird-eye view with an estimated Homography. In this way, the transformed view will have a uniform pixel resolution for the real-world longitude and latitude coordinate. 

% \begin{figure}
%     \centering{\includegraphics[width=0.9\linewidth]{images/calibration.jpg}} \\
%     \caption{We transform a set of ground planar surfaces from a real roadside camera (\#1) to a virtual one (\#2) with uniform geographic resolution. Then, pixel-wise longitude/latitude masks can be generated with the estimated Homography and camera intrinsic parameters.}
%     \label{fig:calibration}
% \end{figure}

The area for perception is represented by a piece-wise segmented planar surface. We manually select a set of ground landmarks (e.g., pavement or roadside static objects) and annotate their pixel coordinate as well as real-world coordinate with Google Maps. For each segment, an Homography matrix $\bm{H}$ can be easily estimated with least square regression and RANSAC consensus between the two groups of landmark sets. A longitude mask $M_{lon}$ and a latitude mask $M_{lat}$ thus can be generated by projecting each pixel of the camera view to the real-world coordinate. Given the pixel location of any detected objects, their localization can be easily retrieved 
from lookup tables:
\begin{equation}
    (x, y) = (M^{(1,..,P)}_{lon}(i,j), M^{(1,..,P)}_{lat}(i,j)),
\end{equation}
where $(i,j)$ is the bottom center pixel coordinate of an object and $(x, y)$ is the estimated longitude and latitude value. $P$ is the number of segmented planers.

The proposed solution also applies to fish-eye cameras. We assume the camera lens follow a generic radially symmetric model~\cite{kannala2006generic} $r(\theta)=k_1\theta + k_2\theta^3 + k_3\theta^5 + \dots$. With the landmark pairs, the camera intrinsic matrix $\bm{K}$ and the distortion coefficients $d_i$ can be estimated~\cite{shah1996intrinsic}. Then, by back-transforming the landmark points to an undistorted camera view, the Homography $\bm{H}^{(1,..P)}$ and the longitude/latitude masks can be generated in a way similar to pinhole cameras. 
% Fig.~\ref{fig:calibration} shows the calibration process.

\subsection{Object Tracking and Information Fusion}

The object tracker is built on top of SORT (Simple Online and Realtime Tracking)~\cite{bewley2016simple}, a popular online object tracking method. The basic idea is using a Kalman Filter~\cite{kalman1960new} and the Hungarian Algorithm~\cite{kuhn1955hungarian} for object state prediction and box matching. Instead of using pixel coordinates, we found using the world coordinate can better deal with camera distortions, especially when tested on fisheye cameras. The state of the Kalman Filter is defined as follows: 
\begin{equation}
    \bm{x} = [x_c, y_c, s, r, v_x, v_y, v_s, v_r]^T,
\end{equation}
where ($x_c, y_c$) are the location of the object; $s$ and $r$ are the area and aspect-ratio of the bounding box; $v_x$, $v_y$, $v_s$ $v_r$ are the derivatives of $x_c$, $y_c$, $s$, and $r$. We set the maximum age of any consecutive un-detected objects to 3.

To fuse the detections from multiple cameras, we divide the map into several regions according to the camera location. The fusion is performed before the tracking, with only those high-certainty detection of each camera being used. Since the tracking is only performed based on the 3D locations, the proposed fusion design makes our system capable of tracking cross-camera moving objects with consistent identities.

\begin{figure}
    \centering{\includegraphics[width=0.9\linewidth]{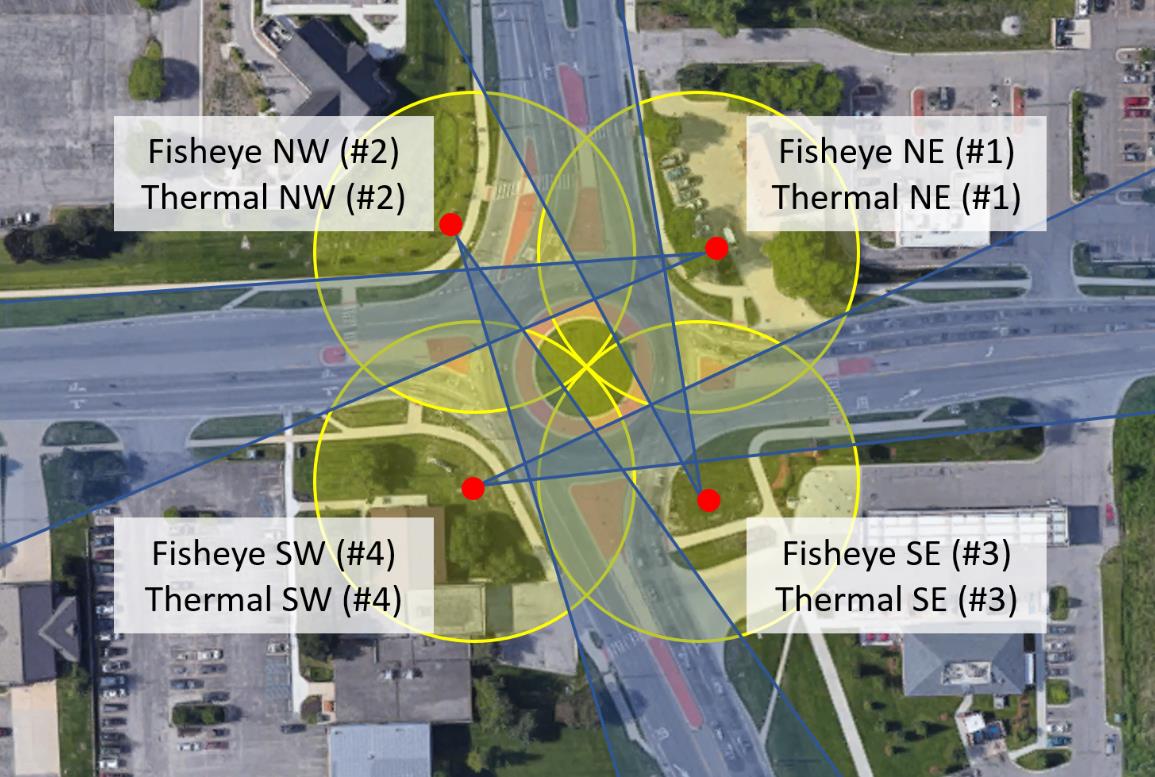}} \\
    \caption{Our real-world test environment: placement of four 360\degree \ fisheye cameras and four thermal cameras at State St. and W. Ellsworth Rd roundabout, Ann Arbor, MI.}
    \label{fig:camera_placement}
\end{figure}

\begin{figure}
    \centering{\includegraphics[width=0.9\linewidth]{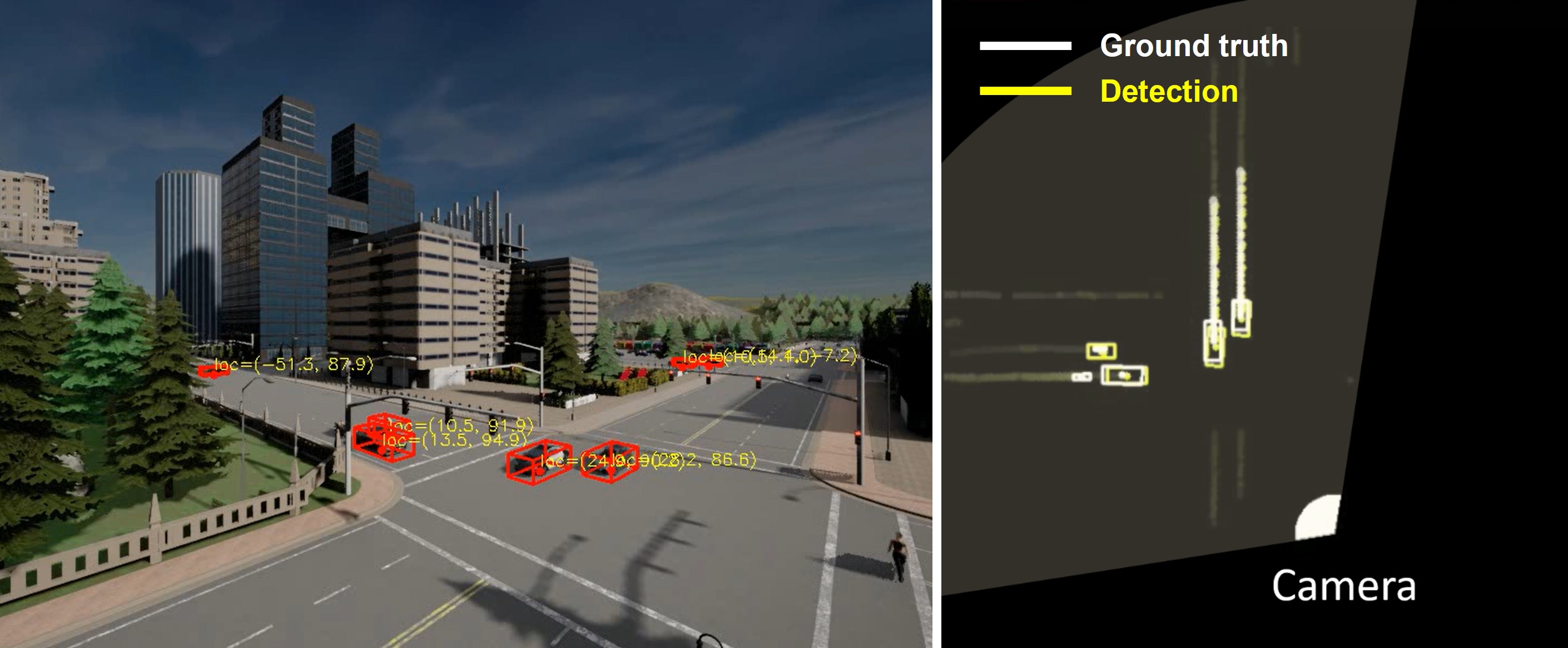}} \\
    \caption{Visualization of object detection, 3D localization, and tracking with a virtual camera placed at CARLA ``Town 05''.}
    \label{fig:carla_det_rst}
\end{figure}

\section{Experimental Results and Analysis}

\subsection{Experimental Setup}

We evaluate our method in both simulation and real-world traffic environments:

\subsubsection{Simulation Environment}

We generate our synthetic dataset with CARLA Simulator~\cite{Dosovitskiy17}. We place four cameras at four corners of an intersection of CARLA ``Town 05''. For each camera, 16 video clips are collected, with $4\times16\times1000$ frames in total. Video \#1 - \#15 are used for training and video \#16 is used for evaluation. We randomly generate 100 vehicles in each video clip. 3D bounding boxes of vehicles in both pixel coordinate and real-world coordinate are recorded. The clock rate is set to 2fps for training and 10fps for testing.

\subsubsection{Real-world Environment}
We evaluate our framework at a roundabout located at the intersection of W Ellsworth Rd and State St. in Ann Arbor, MI, with two groups of cameras - four 360 degree fisheye cameras and four long-range thermal cameras. The cameras are placed at the four corners of the roundabout. For each camera, we annotated 1000 images, with 90\% for training and 10\% for testing. The bottom rectangle of each vehicle is annotated. The annotation of all images took 400 man-hours in total. Fig.~\ref{fig:camera_placement} shows the placement of the cameras.

\subsubsection{Training Details}

We use MobileNet-v2~\cite{sandler2018mobilenetv2} as the backbone of our detector. The detector is trained for 100 epochs using the Adam optimizer with $\text{batch\_size}$=$16$ and $\text{learning\_rate}$=$0.0005$. 
We set $\beta_1$=$\beta_2$=$\beta_3$=$0.01$. When training on the roundabout data, we ignore the vehicle height and predict 2D boxes in pixel size since we do not have their 3D ground truth. Training data augmentation is performed with random image clipping, random gamma correction, and random jittering. The image color is removed at the input of the detector for better adapting to seasonal changes. 

\begin{figure}
    \centering{\includegraphics[width=0.9\linewidth]{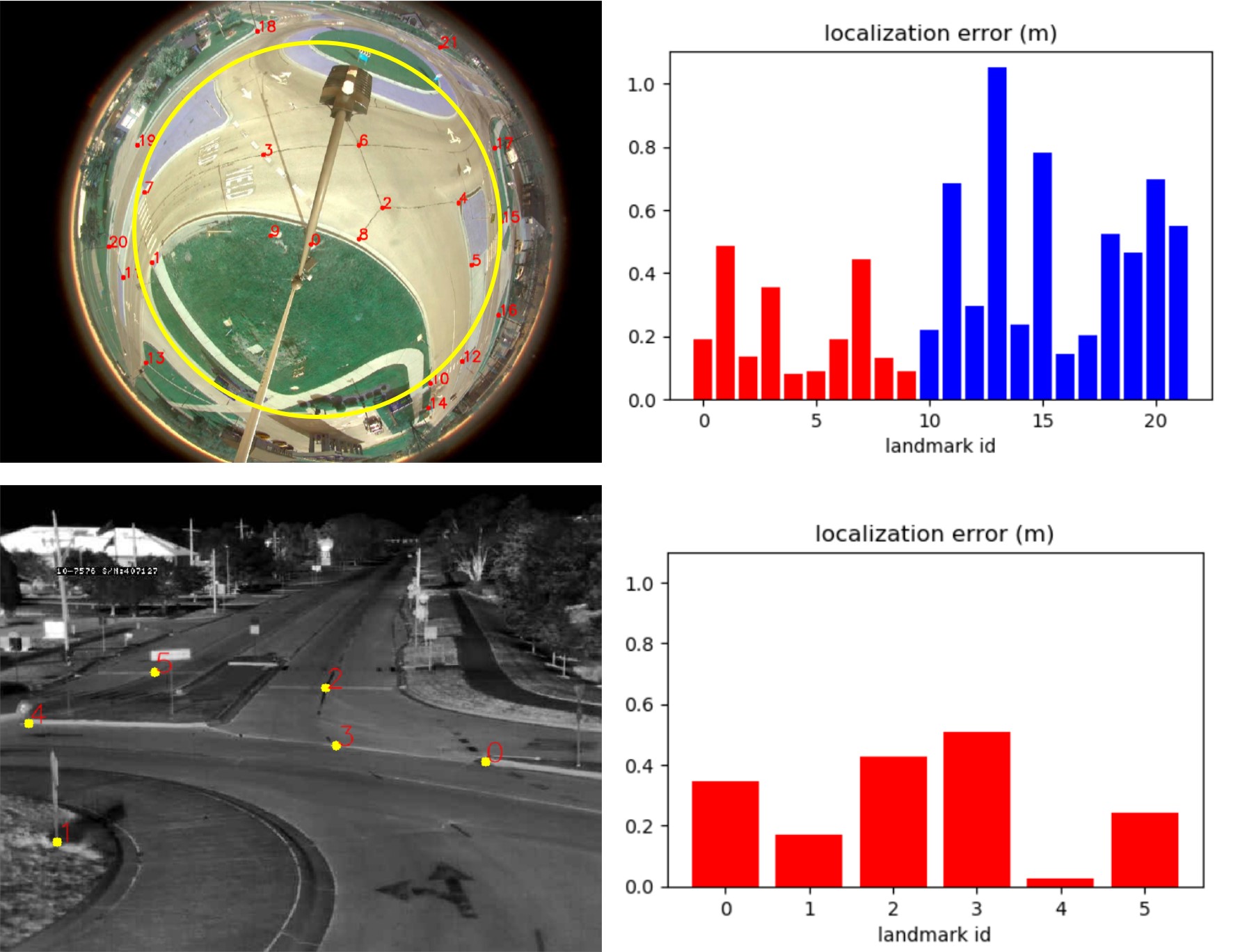}} \\
    \caption{Landmarks and their localization error (m). The red bars correspond to those in-ROI landmarks.}
    \label{fig:calib_error}
\end{figure}

\begin{figure}
    \centering{\includegraphics[width=0.9\linewidth]{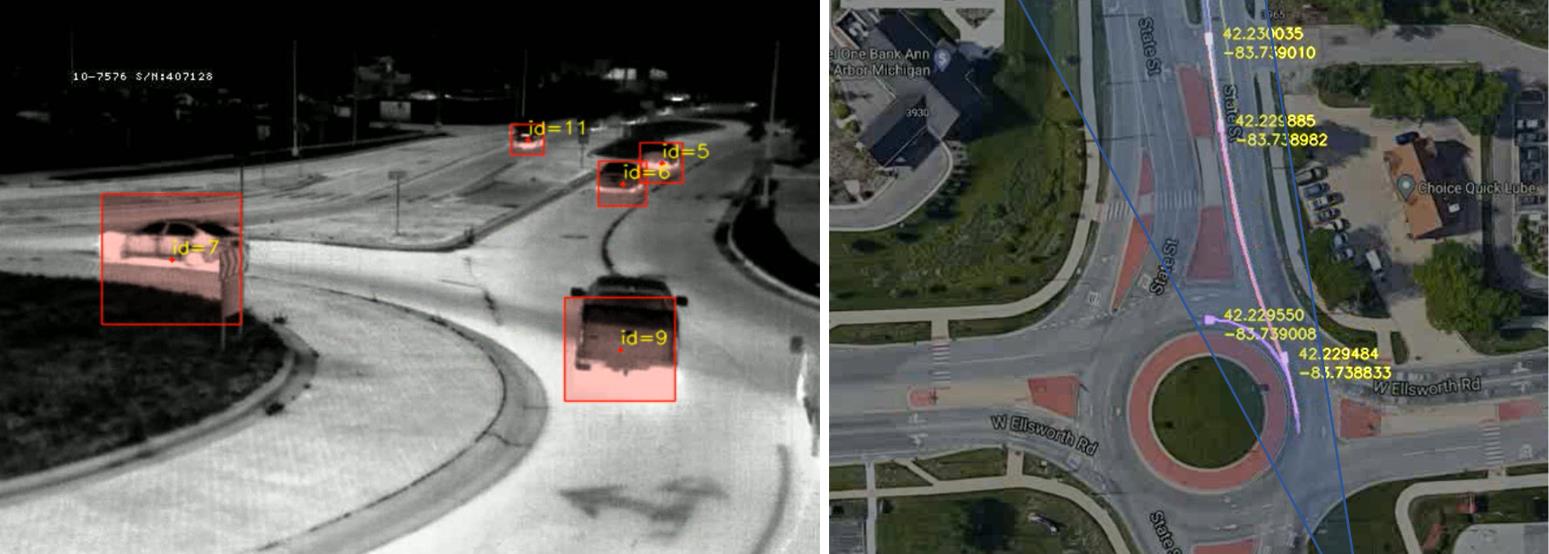}} \\
    \caption{Visualization of detection, localization, and tracking results with roundabout long-range thermal cameras.}
    \label{fig:flir_det_rst}
\end{figure}

\begin{figure}
    \centering{\includegraphics[width=0.9\linewidth]{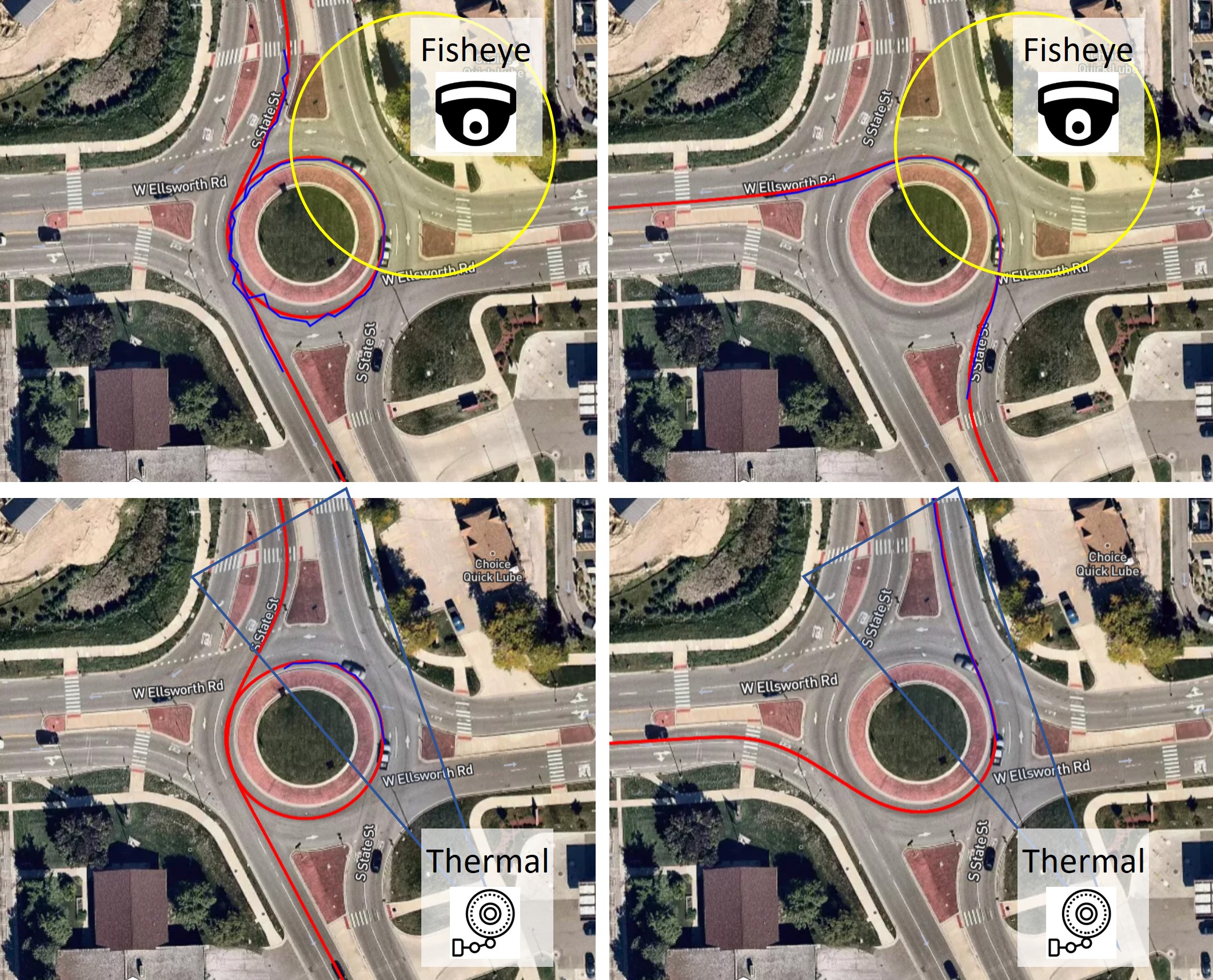}} \\
    \caption{Detected Lincoln MKZ trajectories (blue curve) and the trajectories recorded by RTK as ground truth (red curve). The above two images show the results from fish-eye cam \#1 and the bottom two are from thermal cam \#3. See Fig.~\ref{fig:camera_placement} for the camera placement.}
    \label{fig:mkz_trajectory}
\end{figure}

\begin{table}\small
\begin{center}
\begin{tabular}{l|ccc}
\toprule
           & CARLA & AA-Fisheye & AA-Thermal \\ 
\midrule
w/o TopK select loss & \multicolumn{3}{c}{--- fail to converge ---} \\
full implementation & 0.880 & 0.912 & 0.886 \\
\bottomrule
\end{tabular}%
\end{center}
\vspace{-0.5em}
\caption{2D box-level accuracy (AP) of our detector on different sensors. The results are directly from the detection of individual images. No cross-camera fusion is adopted.}
\label{tab:map}%
\end{table}%

\begin{table}\small
\begin{center}
\begin{tabular}{l|ccc}
\toprule
           & Loc Err & Yaw Err & 3D Size Err \\ 
\midrule
w/o TopK select loss & \multicolumn{3}{c}{--- fail to converge ---}  \\
w/o bottom-center pred & 3.210 (m) & 9.512\degree & 0.448 (m)  \\
full implementation & 0.984 (m) & 9.510\degree & 0.451 (m) \\
\bottomrule
\end{tabular}%
\end{center}
\vspace{-0.5em}
\caption{Pose and 3D size estimation error of our method with different configurations on the CARLA dataset.}
\label{tab:pose_size_loc}%
\end{table}%

\subsection{Bottom-center-aware Detection}

The accuracy of the detector is evaluated on both synthetic images and real images. We follow VOC07~\cite{everingham2010pascal} detection metrics and calculate the mean average precision on different datasets. The VOC box-iou threshold is set to 0.5. Other thresholds are not reported here as bounding-box localization is not the focus of this paper. In~\ref{sec:loc}, we will conduct a more detailed evaluation of 3D localization accuracy.

In Table~\ref{tab:map} and Table~\ref{tab:pose_size_loc}, we show box-level detection accuracy (VOC avg precision) and pose/3d-size error, respectively. In Fig.~\ref{fig:carla_det_rst}, Fig.~\ref{fig:teaser}, and Fig.~\ref{fig:flir_det_rst}, we show the detection + localization result with CARLA images, fish-eye images, and thermal images. When calculating the pose and size error, we only take into account those successful detections. Since we do not have the ground-truth of vehicle pose/size from real-world images, we only evaluate this part in CARLA simulation. An ablation study is also conducted where we remove the TopK selection in Eq.~\ref{eq:bottom-center-loss}, and replace the bottom center prediction with 2D box center prediction. The top rows of Table~\ref{tab:map} and Table~\ref{tab:pose_size_loc} shows the ablation results. Observe when removing the Top-K selection, the training fails to converge. Also, replacing the bottom center prediction with a conventional 2D center prediction caused a noticeable decrease in the localization accuracy.

\subsection{Localization}
\label{sec:loc}

In this experiment, the calibration and end-to-end localization error are evaluated for both fisheye and thermal images.

\subsubsection{Calibration Error Analysis}

Every camera equipped at the roundabout is calibrated manually with 5-20 landmarks labeled on Google Maps. We set the number of segmented planer to one for pinhole camera and four for fisheye camera. We compare the longitude/latitude lookup values at the landmark locations with their ground truth. Fig.~\ref{fig:calib_error} shows the landmark distribution and their localization errors. Since we mainly care about the area underneath the camera (distant area can be covered by other cameras), we divide the map region into two groups: ``region of interest (in-ROI)'' and `` out of the region of interest (out-ROI)''. For a fish-eye camera, we define its ROI as a circular area centered at the camera location with a radius of 25 meters while for a long-range thermal camera, we define its ROI as the $<$200m area within its field of view. Fig.~\ref{fig:calib_error} shows the calibration error. For fisheye cameras, the average in-ROI error (within the yellow circle, marked as red in the bar-plot) is 0.219 $\pm$ 0.145 m. The out-ROI error (marked as blue in the bar-plot) is 0.489 $\pm$ 0.268 m. For thermal cameras, the error is 0.288 $\pm$ 0.162 m.

\begin{table}\small
\begin{center}
\begin{tabular}{l|ccc}
\toprule
 & \multicolumn{2}{c}{Fisheye} & Thermal \\
     & In-ROI (m) & Out-ROI (m) & In-ROI (m)\\ 
\midrule
Trip \#1 & 0.478 $\pm$ 0.248 & 1.038 $\pm$ 0.965  & 0.615 $\pm$ 0.340 \\
Trip \#2 & 0.377 $\pm$ 0.218 & 0.779 $\pm$ 0.747  & 0.339 $\pm$ 0.251  \\
Trip \#3 & 0.408 $\pm$ 0.167 & 1.154 $\pm$ 1.138  & 1.334 $\pm$ 0.792  \\
Trip \#4 & 0.217 $\pm$ 0.162 & 0.596 $\pm$ 0.559  & 0.373 $\pm$ 0.363  \\
Trip \#5 & 0.368 $\pm$ 0.195 & 0.969 $\pm$ 0.849  & 0.860 $\pm$ 0.622 \\
Trip \#6 & 0.401 $\pm$ 0.210 & 1.491 $\pm$ 2.031  & 0.837 $\pm$ 0.456 \\
\midrule
All & 0.377 $\pm$ 0.207 & 0.964 $\pm$ 1.085  & 0.820 $\pm$ 0.546 \\
\bottomrule
\end{tabular}%
\end{center}
\vspace{-0.5em}
\caption{Trajectory error between the detection and the ground truth (RTK) with fisheye and thermal cameras.}
\label{tab:mkz_trajectory}%
\end{table}%

\begin{figure}
    \centering{\includegraphics[width=0.9\linewidth]{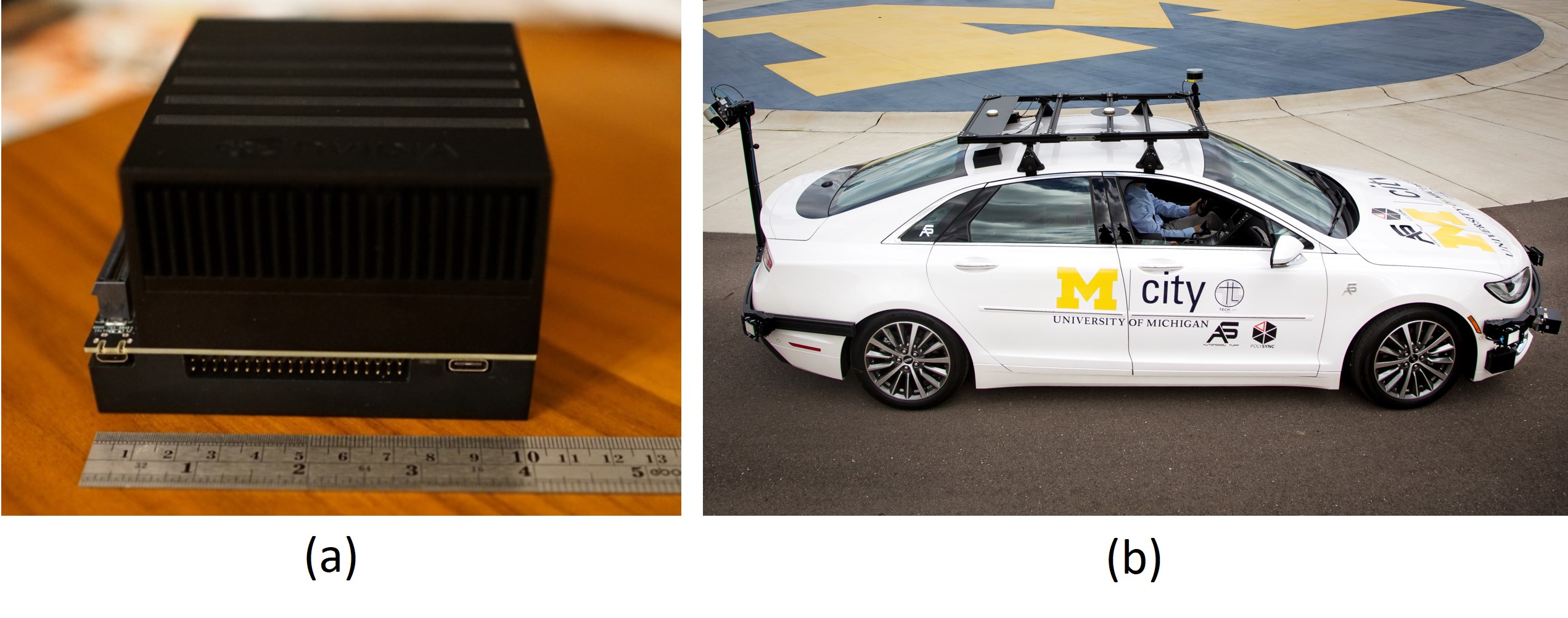}} \\
    \caption{(a) The edge device where our method is deployed and tested. (b) Our testing platform - a Hybrid Lincoln MKZ equipped with a high-precision RTK~\cite{xu2018accurate, xu2019design}.}
    \label{fig:edge_device}
\end{figure}

\subsubsection{Evaluation with Connected Vehicle}

We deploy our system on an edge device (Jetson AGX XAVIER) at the City of Ann Arbor and provide 7x24 monitoring service of the roundabout traffic. A connected automated vehicle\footnote{\url{https://mcity.umich.edu/}} — a Hybrid Lincoln MKZ~\cite{xu2018accurate, xu2019design} equipped with a high-precision RTK and an Inertial Measurement Unit (IMU), is used to test our system. With the vehicle and sensors, we can measure the vehicle location in real-time. The vehicle and the edge device are shown in Fig.~\ref{fig:edge_device}.

\begin{figure}
    \centering{\includegraphics[width=0.9\linewidth]{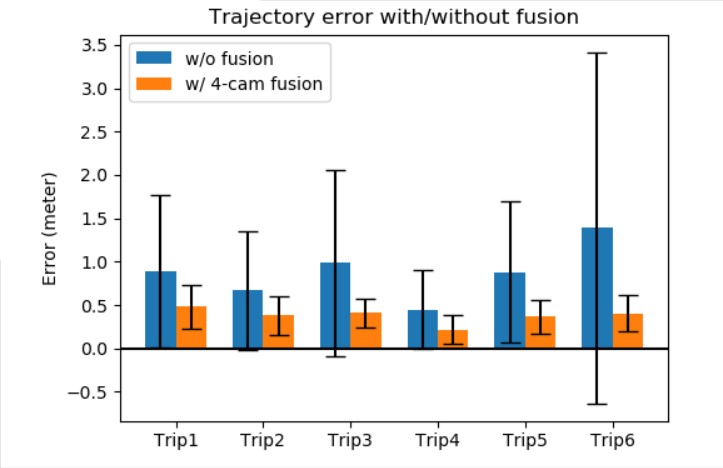}} \\
    \caption{Trajectory error with or without the fusion of the four fisheye cameras.}
    \label{fig:fusion_err_bar}
\end{figure}

The vehicle is driven through the roundabout six times in two separate days: June 30th, 2021, and July 19th, 2021, recording the trajectories by RTK GPS as the ground truth. Fig.~\ref{fig:mkz_trajectory} shows the detected trajectories alongside with the ground-truth. Table~\ref{tab:mkz_trajectory} shows the localization error. For each trip, the error is calculated as the average project distance between the localization points and the ground truth trajectory. The average In-ROI error over 6 trials for fish-eye and thermal cameras are  0.377 m and 0.820 m respectively. Fig.~\ref{fig:fusion_err_bar} shows the localization error within the entire roundabout area before and after the fusion of all four fisheye cameras. With fusion, the average localization error is reduced from 0.834 m $\pm$ 1.037 m to 0.377 m $\pm$ 0.207 m. The fusion can therefore greatly improve both the localization accuracy and stability. Note that the large variance of the 6th trip error is caused by the camera shake in the wind. Nevertheless, we choose to report this non-ideal trip and include it in the performance analysis to give an end-to-end accuracy considering all practical issues.

\begin{table}\small
\begin{center}
\begin{tabular}{l|cc}
\toprule
     & 1-way processing & 4-way processing \\ 
\midrule
I7-9700K + 2070S & 160 fps / 6 ms & 60$\times$4 fps / 16 ms \\
I7-8750H + 1050Ti & 60 fps / 16 ms  & 16$\times$4 fps / 60 ms \\
Jetson AGX XAVIER & 50 fps / 20 ms & 18$\times$4 fps / 55 ms \\
\bottomrule
\end{tabular}%
\end{center}
\vspace{-0.5em}
\caption{The speed performance of (frames per second and delay) of the proposed framework on different devices.}
\label{tab:speed}%
\end{table}%

\subsection{Speed performance}

We test the inference speed of our framework on multiple platforms with different computational capabilities. Table~\ref{tab:speed} shows the detailed speed performance of our system. With half-precision inference speedup, the whole processing pipeline of our system (detection + localization + fusion + tracking) achieves 160fps on an I7-9700K+2070S desktop and 50fps on a Jetson AGX XAVIER edge device. When handling 4-way input video streams simultaneously, our system still achieves real-time processing speed, with $60\times 4$ fps and $18\times 4$ fps on the two platforms respectively.

\section{CONCLUSIONS}

We propose a vision-based traffic scene perception framework with object detection, localization, tracking, and sensor fusion. Owing to the decoupling design, the framework can be trained solely based on 2D annotations, which greatly overcomes difficulties in field deployment and migration. We tested our system with both real-world connected and automated vehicles and simulation environment, and achieve 0.4-meter localization accuracy within an entire 100x100 m$^2$ two-lane roundabout area. The all-components end-to-end perception delay is less than 20ms. The proposed method provides a novel solution for practical roadside perception and shows great potential in the cooperative perception of automated vehicles with infrastructure support.

\section*{ACKNOWLEDGMENT}

This work was partially supported by Mcity and the College of Engineering of the University of Michigan and the Ford Motor Company.

%%%%%%%%%%%%%%%%%%%%%%%%%%%%%%%%%%%%%%%%%%%%%%%%%%%%%%%%%%%%%%%%%%%%%%%%%%%%%%%%

\bibliographystyle{IEEEtran}
\bibliography{IEEEabrv, egbib}

\end{document}